\documentclass[conference]{IEEEtran}
\IEEEoverridecommandlockouts

\usepackage{cite}
\usepackage{amsmath,amssymb,amsfonts}
\usepackage{algorithmic}
\usepackage{graphicx}
\usepackage{xcolor}
\usepackage{textcomp}
\usepackage{float}

\usepackage[font=small,skip=2pt]{caption}
\usepackage{subcaption}

\def\BibTeX{{\rm B\kern-.05em{\sc i\kern-.025em b}\kern-.08em
  T\kern-.1667em\lower.7ex\hbox{E}\kern-.125emX}}

\makeatletter
\newcommand{\linebreakand}{%
  \end{@IEEEauthorhalign}
  \hfill\mbox{}\par
  \mbox{}\hfill\begin{@IEEEauthorhalign}
}
\makeatother

\begin{document}

\title{
LiGen: GAN-Augmented Spectral Fingerprinting for Indoor Positioning
}

\author{\IEEEauthorblockN{
Jie Lin\textsuperscript{*}\IEEEauthorrefmark{2},
Hsun-Yu Lee\textsuperscript{*}\IEEEauthorrefmark{2},
Ho-Ming Li\IEEEauthorrefmark{3},
and Fang-Jing Wu\IEEEauthorrefmark{4}}
\IEEEauthorblockA{
\IEEEauthorrefmark{2}Department of Information Management, National Taiwan University, Taiwan\\
\IEEEauthorrefmark{3}Graduate Institute of Networking and Multimedia, National Taiwan University, Taiwan\\
\IEEEauthorrefmark{4}Department of Computer Science and Information Engineering, National Taiwan University, Taiwan\\
Email: b11705048@ntu.edu.tw,\ b11705022@ntu.edu.tw,\ r12944054@csie.ntu.edu.tw,\ fangjing@csie.ntu.edu.tw}
}

\maketitle

\begingroup
\renewcommand\thefootnote{\textsuperscript{*}}
\footnotetext{These authors contributed equally to this work.}
\endgroup

\begin{abstract}
Accurate and robust indoor localization is critical for smart building applications, yet existing Wi-Fi-based systems are often vulnerable to environmental conditions. This work presents a novel indoor localization system, called \textbf{LiGen}, that leverages the spectral intensity patterns of ambient light as fingerprints, offering a more stable and infrastructure-free alternative to radio signals. To address the limited spectral data, we design a data augmentation framework based on generative adversarial networks (GANs), featuring two variants: \textbf{PointGAN} and \textbf{FreeGAN}. Our positioning model, leveraging a Multi-Layer Perceptron (MLP) architecture to train on synthesized data, achieves submeter-level accuracy, outperforming Wi-Fi-based baselines by over 50\%. LiGen also demonstrates strong robustness in cluttered environments. To the best of our knowledge, this is the first system to combine spectral fingerprints with GAN-based data augmentation for indoor localization.
\end{abstract}


\section{Introduction}
Submeter-accurate indoor localization underpins numerous applications, yet GPS inaccessibility and intricate propagation dynamics pose formidable challenges~\cite{Grosswindhager2018SALMA}. Wi-Fi RSSI, though prevalent, suffers from instability owing to multipath fading, obstructions, and interference, yielding erratic performance in cluttered settings~\cite{Ahmad2024,Singh2021WiFiML,Sadowski2018RSSILocalization}.

To mitigate these shortcomings, nascent studies explore optical signals as viable alternatives or adjuncts for indoor positioning. This avenue, however, remains underexamined. Herein, we investigate spectral light intensity patterns---ambient light signatures acquired through off-the-shelf spectral sensors---as resilient and discriminative localization fingerprints. In contrast to Wi-Fi, these signals necessitate no bespoke infrastructure and demonstrate diminished susceptibility to environmental perturbations, rendering them a compelling, economical, and infrastructure-agnostic paradigm for indoor navigation~\cite{Zhang2016LiTell, Ashraf2020Smartphone}.

We propose \textbf{LiGen}, a novel indoor localization system that integrates spectral fingerprinting with Generative Adversarial Networks (GANs) \cite{goodfellow2014generativeadversarialnetworks} for data augmentation, enhancing model robustness against data scarcity. Data scarcity in spectral-based localization stems from the labor-intensive process of acquiring dense, spatially varied fingerprints, which demands extensive manual surveys and is prone to inconsistencies from diurnal lighting fluctuations or occlusions. By synthesizing realistic augmentations, LiGen bridges this gap, enabling effective training with sparse real-world data. The contributions of this work are threefold:
\begin{figure}[t]
    \centering
    \includegraphics[width=0.47\textwidth]{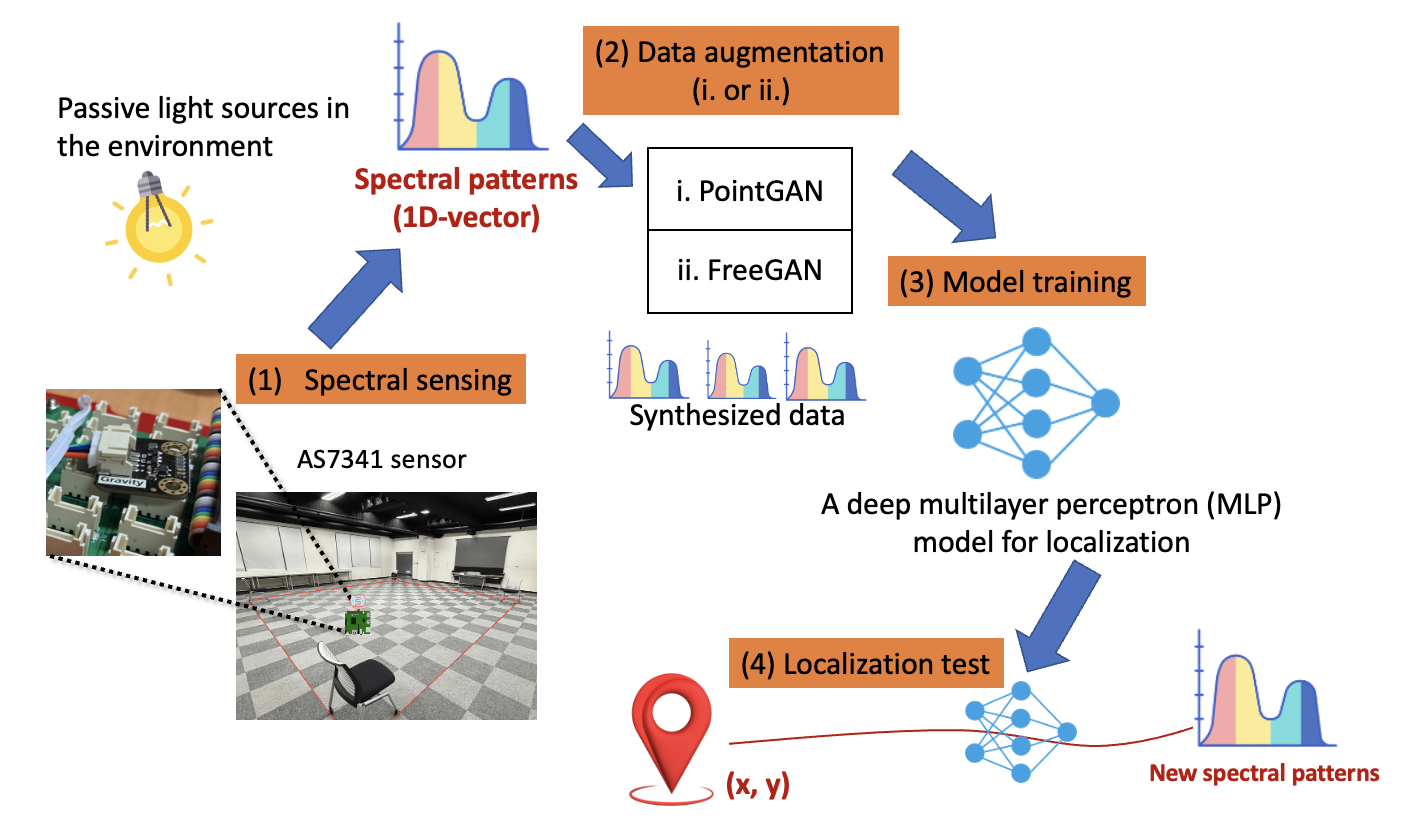}
    \caption{The system overview of LiGen.}
    \label{fig:system_structure}
\end{figure}
\begin{itemize}
    \item We show that ambient spectral fingerprints yield higher coordinate-level accuracy than Wi‑Fi RSSI, providing a more robust, infrastructure‑free modality for indoor localization.
    \item We develop a GAN-based augmentation pipeline tailored to spectral fingerprints—PointGAN (coordinate conditioned) and FreeGAN (unconditioned with WLM pseudo‑labels)—and train a compact MLP that leverages synthetic data to improve performance under data scarcity.
    \item We validate LiGen in clean and cluttered settings, demonstrating that light‑based fingerprints are less sensitive to occlusions and multipath, sustaining superior accuracy where RSSI methods degrade.
\end{itemize}

To realize these contributions, LiGen introduces two GAN variants: \textbf{PointGAN}, a conditional GAN that generates fingerprints based on spatial coordinates, and \textbf{FreeGAN}, which generates unconditioned samples and leverages a weak localization model for pseudo-label assignment. Both approaches enrich the training dataset and improve generalization of the localization model, implemented as a deep MLP \cite{rumelhart1986learning}.

To the best of our knowledge, LiGen is the first system to integrate ambient light spectral information and GAN-based data augmentation for submeter-level indoor localization. Our extensive experiments validate the effectiveness of this approach, highlighting its potential for robust, real-world deployment.


\section{Related Work}
Indoor localization has been extensively studied as GPS is ineffective indoors, motivating alternatives based on electromagnetic (EM) signals, optical sensing, and machine learning. Early efforts largely used Wi-Fi RSSI, e.g., DataLoc+~\cite{Hilal2021-DataLoc}, which applied data augmentation to improve room-level accuracy. However, RSSI methods are highly sensitive to obstacles, multipath, and interference, leading to inconsistent performance~\cite{Ahmad2024, Yang2013, Singh2021WiFiML, Sadowski2018RSSILocalization}.

Light-based approaches emerged as alternatives. LuxTrace~\cite{Randall2007-LuxTrace} and Pulsar~\cite{Zhang2017Pulsar} demonstrated positioning using light intensity, while Wang et al.~\cite{Wang2020-LSTM} fused light and magnetic signals via LSTMs. These relied on coarse intensity, unlike SpectralLoc~\cite{wang2022spectrallocindoorlocalizationusing}, which used the AS7265x sensor to capture ambient spectral patterns and achieved submeter accuracy with a CNN–attention model. Inspired by SpectralLoc, LiGen leverages spectral fingerprints from the AS7341 sensor~\cite{amsAS7341UserGuide} and enhances training with GAN-based augmentation.

Deep learning is central to localization~\cite{Chen2024-Survey}, with models such as MLPs capturing complex sensor–space relationships. Data augmentation is equally critical: while DataLoc+ targets Wi-Fi RSSI, GANs have shown broader success in generating realistic signals~\cite{Shi2019-GAN-Spoofing}. Building on this, LiGen introduces PointGAN and FreeGAN to synthesize spectral fingerprints, addressing data scarcity. To our knowledge, it is the first system applying GAN-based augmentation to spectral light data for indoor localization.

In summary, LiGen integrates SpectralLoc’s spectral fingerprinting~\cite{wang2022spectrallocindoorlocalizationusing}, DataLoc+’s augmentation~\cite{Hilal2021-DataLoc}, and GAN-based synthesis~\cite{Shi2019-GAN-Spoofing}, advancing light-based localization~\cite{Wang2020-LSTM, Randall2007-LuxTrace} with deep learning~\cite{Chen2024-Survey} for high-accuracy indoor positioning.

\section{System Architecture}
\subsection{System Overview}


LiGen achieves coordinate-level indoor localization using ambient-light spectral fingerprints, which are more robust than Wi-Fi RSSI. Its pipeline includes:
\begin{enumerate}
\item \textbf{Sensing:} AS7341 sensor captures 10-channel spectral fingerprints.
\item \textbf{Augmentation:} PointGAN (location-conditioned) and FreeGAN (pseudo-labeled) generate synthetic fingerprints~\cite{ho2020denoisingdiffusionprobabilisticmodels}. GANs are particularly suited for this task as they train efficiently compared to diffusion or autoregressive models, require no complex sampling procedures, and directly capture data distributions, making them fast and practical for large-scale augmentation.
\item \textbf{Localization:} A simple MLP maps fingerprints to 2D coordinates, demonstrating that—when paired with effective augmentation—lightweight models can achieve strong localization performance without complex architectures.
\end{enumerate}

Figure~\ref{fig:system_structure} summarizes the flow: fingerprints are collected, augmented, and passed to the localization model for coordinate prediction.

\subsection{System Model}

We consider an indoor region $\mathcal{R} \subset \mathbb{R}^2$ with $N$ predefined reference locations 
$\mathbf{p}_i = (x_i, y_i)$ for $i = 1, \dots, N$.  
At each location $\mathbf{p}_i$, the spectral sensor collects $M_i$ ambient-light fingerprints, denoted as 
$\{\mathbf{x}_i^{(k)}\}_{k=1}^{M_i}$. Each $\mathbf{x}_i^{(k)}$ is a 10-dimensional fingerprint representing spectral intensities across different channels, defined as
\begin{small}
\begin{IEEEeqnarray}{rCl}
\mathbf{x}_i^{(k)} 
&=& \{(F1, I^{(1)}_i[k]), \dots, (F8, I^{(8)}_i[k]), \nonumber\\
&& (\mathrm{NIR}, I^{(9)}_i[k]), (\mathrm{Clear}, I^{(10)}_i[k])\}. \label{eq:vector}
\end{IEEEeqnarray}
\end{small}

Here, $I^{(j)}_i[k]$ is the intensity measured at the $j$-th spectral channel for the $k$-th sample at location $\mathbf{p}_i$.
The AS7341 sensor defines fixed spectral bands: F1–F8 span visible light from 415 to 680nm, NIR captures near-infrared, and Clear measures total visible intensity (400–700,nm).

The overall dataset is defined as
\[
\mathcal{D} = \bigcup_{i=1}^N \big\{ (\mathbf{x}_i^{(k)}, \mathbf{p}_i) \,|\, k = 1, \dots, M_i \big\}.
\]

To address data scarcity, PointGAN $G_P(\mathbf{z}, \mathbf{c})$ synthesizes location-conditioned fingerprints, while FreeGAN $G_F(\mathbf{z})$ produces unlabeled samples pseudo-labeled by a weak model $f_\phi$.
The augmented dataset $\mathcal{D}_{\mathrm{aug}}$ is then used to train the final localization model $f_\theta : \mathbb{R}^{10} \rightarrow \mathbb{R}^2$.

\section{Methodology}
\subsection{Localization Model}

\begin{figure}[ht]
    \centering
    \includegraphics[width=0.47\textwidth]{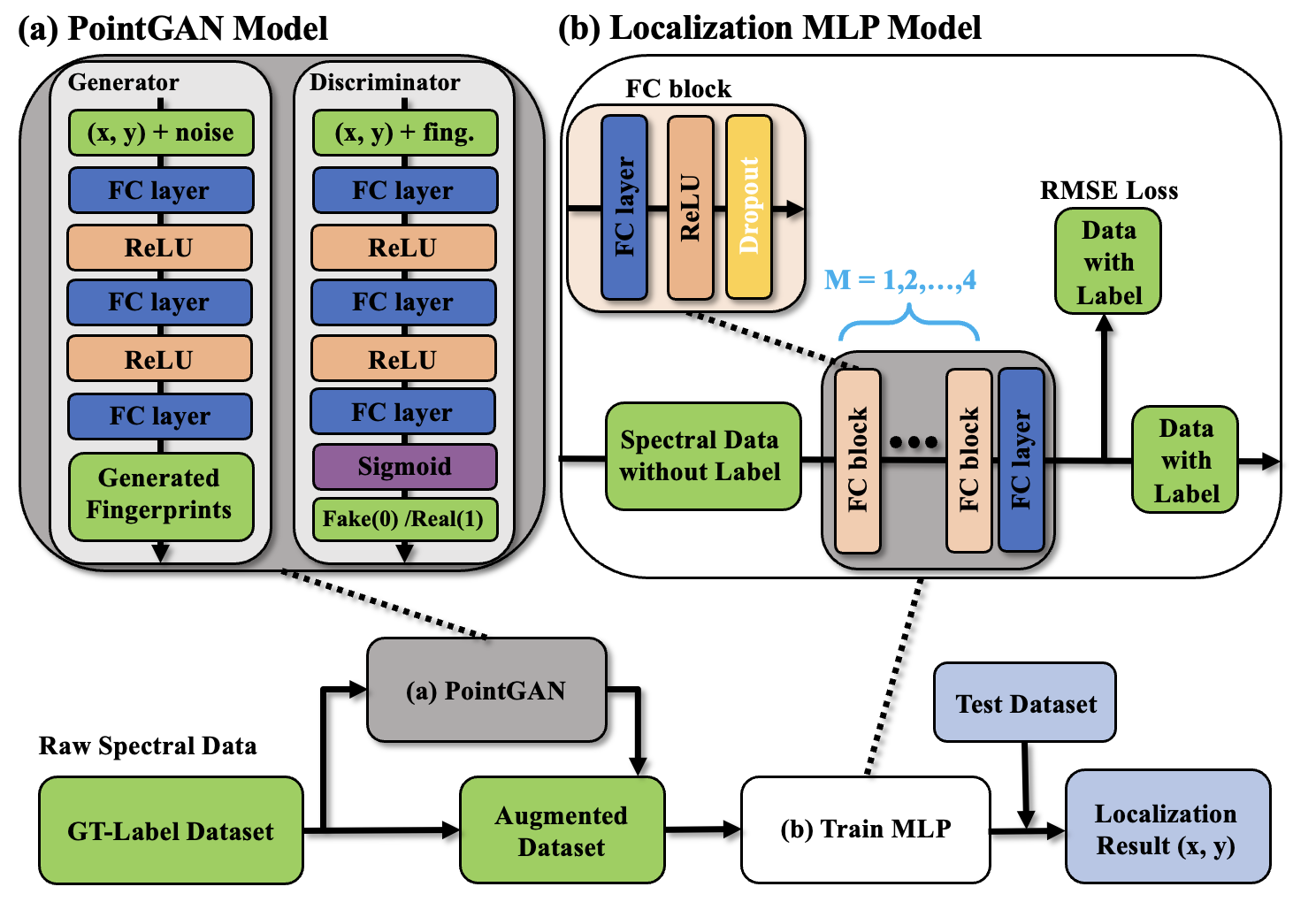}
    \caption{The system design of LiGen using PointGAN for fingerprint augmentation.}
    \label{fig:ligen_model}
\end{figure}

We employ a deep learning-based localization model to map fingerprints to 2D positions. The model is a Multi-Layer Perceptron (MLP)~\cite{rumelhart1986learning} (Fig.~\ref{fig:ligen_model}b) with:
\begin{itemize}
\item \textbf{Input:} fingerprint features of dimension $d$ (e.g., spectral intensities or Wi-Fi RSSI).
\item \textbf{Hidden layers:} fully connected layers with ReLU~\cite{RELU} activations and dropout~\cite{dropout} for non-linear feature learning and regularization.
\item \textbf{Output:} a 2D coordinate $\hat{\mathbf{y}} = (x, y)$.
\end{itemize}

The model is trained by minimizing the mean squared error (MSE) between the predicted coordinates and the ground truth:
$\mathcal{L}_{\mathrm{MSE}} = \frac{1}{N}\sum_{i=1}^{N} \left\|\hat{\mathbf{y}}_i - \mathbf{y}_i\right\|_2^2$.
During evaluation, we report the root-mean-square error (RMSE) in meters:
$
\mathrm{RMSE} = \sqrt{\mathcal{L}_{\mathrm{MSE}}}.
$

\begin{enumerate}
\item \textbf{Strong Model} $f_\theta$: trained on the augmented dataset $\mathcal{D}_{\mathrm{aug}}$ and used for final localization.
\item \textbf{Weak Model (WLM)} $f_\phi$: trained only on the real dataset $\mathcal{D}$ to provide pseudo-labels for FreeGAN-generated fingerprints, enriching $\mathcal{D}_{\mathrm{aug}}$.
\end{enumerate}

\subsection{Data Augmentation with Generative Adversarial Networks}
 We propose two GAN-based frameworks to augment indoor localization data: PointGAN and FreeGAN. While both methods generate synthetic fingerprints to enhance the training dataset, they differ in their conditioning strategies and data augmentation pipelines.

\subsubsection{PointGAN}
The proposed PointGAN framework adopts a Conditional GAN (CGAN) \cite{mirza2014conditionalgenerativeadversarialnets} architecture to generate synthetic indoor localization fingerprints.

Figure~\ref{fig:ligen_model} illustrates the overall architecture of the proposed LiGen system, which integrates PointGAN for fingerprint augmentation and a deep localization MLP model for coordinate-level localization.

The Generator \(G_P\) takes two inputs:
\begin{itemize}
    \item A random noise vector \(\mathbf{z} \sim \mathcal{N}(0, I)\),
    \item Spatial coordinates \(\mathbf{c} = (x, y)\).
\end{itemize}

It outputs a synthetic fingerprint \(\hat{\mathbf{f}} = G_P(\mathbf{z}, \mathbf{c})\), where \(\hat{\mathbf{f}} \in \mathbb{R}^{\text{fingerprint\_dim}}\), using fully connected layers with ReLU activations.

The Discriminator \(D_P\) receives:
\begin{itemize}
    \item Coordinates \(\mathbf{c} = (x, y)\),
    \item A fingerprint \(\mathbf{f}\), either real or generated.
\end{itemize}

It outputs a scalar \(D_P(\mathbf{c}, \mathbf{f})\), representing the probability that \(\mathbf{f}\) is real, via a Multi-Layer Perceptron (MLP) with Sigmoid activation.

The objective function for PointGAN is designed to train two neural networks—the Generator and the Discriminator—in opposition to each other. The Generator tries to produce synthetic fingerprints that resemble real ones for given spatial coordinates, while the Discriminator attempts to distinguish between real and generated fingerprints.

Mathematically, the objective function is:

\begin{small}
\begin{IEEEeqnarray}{rCl}
\min_{G_P} \max_{D_P} \; V(D_P, G_P) &=& \mathbb{E}_{(\mathbf{c}, \mathbf{f})} \left[ \log D_P(\mathbf{c}, \mathbf{f}) \right] \nonumber\\
&& + \mathbb{E}_{\mathbf{z}, \mathbf{c}} \left[ \log \left( 1 - D_P\bigl(\mathbf{c}, G_P(\mathbf{z}, \mathbf{c})\bigr) \right) \right]
\end{IEEEeqnarray}
\label{eq:pointgan_objective}
\end{small}
The objective follows standard CGAN formulation, rewarding correct discrimination and penalizing misclassification.

By conditioning the Generator on spatial coordinates, PointGAN ensures that the synthetic fingerprints align with actual locations, while the Discriminator refines the Generator's outputs through adversarial training.

\begin{figure}[t]
    \centering
\includegraphics[width=0.47\textwidth]{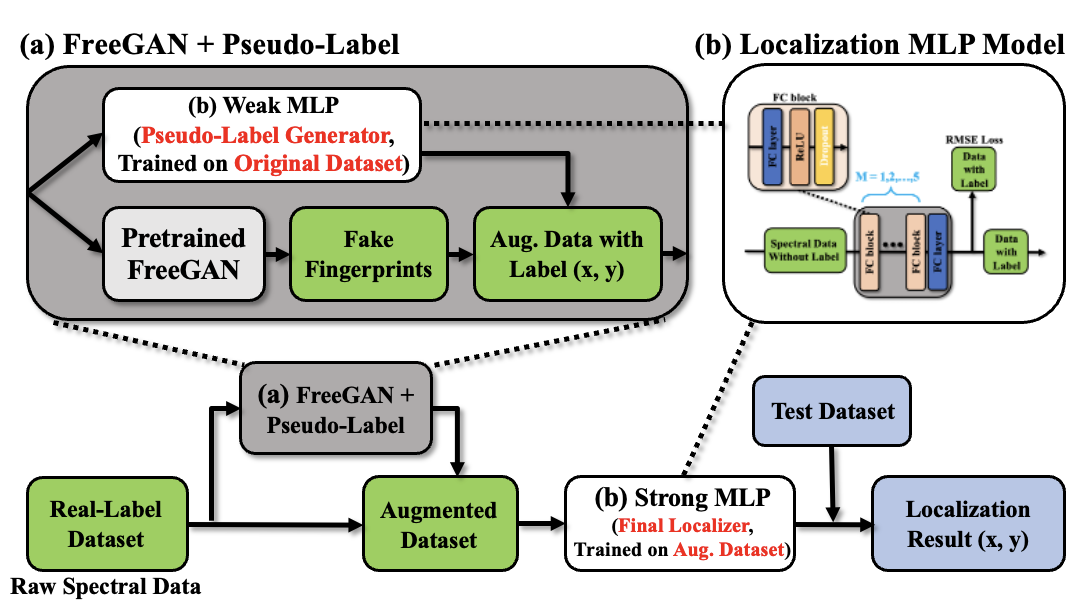}
    \caption{The system design of LiGen using FreeGAN for fingerprint augmentation.}
    \label{fig:freegan_model}
\end{figure}

\subsubsection{FreeGAN}
The proposed FreeGAN framework utilizes an architecture similar to PointGAN but generates synthetic indoor localization fingerprints solely from latent noise—without explicit spatial conditioning. Figure~\ref{fig:freegan_model} illustrates the LiGen system, which integrates FreeGAN for fingerprint augmentation.

\begin{enumerate}
\item \textbf{Weak Model:} Train $f_\phi$ on real data.
\item \textbf{Generation:} The generator $G_F(\mathbf{z}), \mathbf{z}\sim\mathcal{N}(0,I)$ outputs synthetic fingerprints via fully connected layers with ReLU.
\item \textbf{Pseudo-labeling:} The generated fingerprints are passed through the weak localization model $f_\phi$ to obtain pseudo-labels (i.e., estimated spatial coordinates): $ \hat{\mathbf{c}} = f_\phi\bigl(G_F(\mathbf{z})\bigr). $
\item \textbf{Augmentation:} Combine pseudo-labeled samples with real data to form $\mathcal{D}_{\text{aug}}$.
\end{enumerate}

The Generator and Discriminator are trained adversarially to minimize/maximize the following objective, which follows the original GAN formulation~\cite{goodfellow2014generativeadversarialnetworks}:

\begin{small}
\begin{IEEEeqnarray}{rCl}
\min_{G_F} \max_{D_F} \; V(D_F, G_F) &=& \mathbb{E}_{\mathbf{f} \sim p_{\text{data}}(\mathbf{f})} \left[ \log D_F(\mathbf{f}) \right] \nonumber\\
&& + \mathbb{E}_{\mathbf{z} \sim p_{\mathbf{z}}(\mathbf{z})} \left[ \log \left( 1 - D_F\left(G_F(\mathbf{z})\right) \right) \right]
\end{IEEEeqnarray}
\end{small}

where \( p_{\text{data}}(\mathbf{f}) \) denotes the distribution of real fingerprints and \( p_{\mathbf{z}}(\mathbf{z}) \) denotes the noise prior (e.g., \(\mathcal{N}(0, I)\)). Also, the Generator tries to produce synthetic fingerprints that resemble real ones for given spatial coordinates, while the Discriminator attempts to distinguish between real and generated fingerprints.

 The structure of the localization model is shown in detail in Figure~\ref{fig:ligen_model} section (b). Note that there are two Localization MLP models with identical architectures in the pipeline: one is used for assigning pseudo labels, and the other serves as the final model.

By leveraging a weak localization model to provide pseudo-labels, FreeGAN effectively bridges the gap between synthetic data generation and accurate spatial representation, thereby enhancing the robustness of localization models.

\subsection{Training Parameters and Baselines}

All experiments follow a unified setup for fair comparison. Models are trained on a single NVIDIA RTX~A6000 GPU (48\,GB VRAM) with batch size 4096. Optimization uses Adam, with learning rate, dropout, and hidden size tuned by Optuna (Table~\ref{tab:hyperparams}); default parameters are in Table~\ref{tab:default_params}. 

\subsubsection{MLP architecture}
5 fully connected layers: 4 hidden layers with ReLU activations and dropout, followed by a linear 2D output. With the following parameters tuned by Optuna.

\vspace{0.3cm}
\begin{table}[H]
\centering
\caption{Optuna search ranges for MLP hyperparameters.}
\label{tab:hyperparams}
\begin{tabular}{l c}
\hline
Hyperparameter & Search Range \\
\hline
Hidden layer size & 64 – 1024 (log scale) \\
Learning rate     & $10^{-2}$ – $10^{-4}$ (log scale) \\
Dropout rate      & 0.1 – 0.5 (uniform) \\
Trials            & 3 \\
\hline
\end{tabular}
\end{table}

\subsubsection{GAN architectures}
\textbf{PointGAN}: Generator with two hidden layers (128/256, ReLU); discriminator processes coordinate–fingerprint pairs via a similar structure with sigmoid output. Trained 5000 epochs with Adam. \textbf{FreeGAN}: Generator maps noise to fingerprint space via two fully connected layers with batch normalization and LeakyReLU; discriminator mirrors this with LeakyReLU layers and sigmoid output. Trained 5000 epochs with Adam, binary cross-entropy loss, batch 4096.

\vspace{0.3cm}
\begin{table}[H]
\centering
\caption{Default training parameters used across all models.}
\label{tab:default_params}
\begin{tabular}{l c}
\hline
Parameter & Value \\
\hline
\multicolumn{2}{l}{\textbf{General}} \\
Batch size & 4096 \\
Optimizer & Adam \\
\hline
\multicolumn{2}{l}{\textbf{MLP}} \\
Loss & L2 (MSE)\footnotemark \\
Adam $(\beta_1,\beta_2)$ & $(0.9,\ 0.999)$ \\
Epochs & 700 \\
\hline
\multicolumn{2}{l}{\textbf{GANs (PointGAN / FreeGAN)}} \\
Loss & Binary cross-entropy (BCE) \\
Adam $(\beta_1,\beta_2)$ & $(0.5,\ 0.999)$ \\
Epochs & 5000  \\
Learning rate & $10^{-4}$ \\
\hline
\end{tabular}
\end{table}
\footnotetext{Training optimizes MSE; evaluation is reported as RMSE (in meters).}

\subsubsection{Baselines}  
\begin{itemize}
  \item \textbf{DataLoc+ (Wi-Fi)}~\cite{Hilal2021-DataLoc}: Wi-Fi RSSI localization with its original augmentation.
  \item \textbf{DataLoc+ (Spectral)}: Same augmentation applied to spectral fingerprints for fair comparison.
  \item \textbf{MLP (Spectral)}: MLP trained solely on real spectral data, without augmentation.
  \item \textbf{MLP (Wi-Fi)}: MLP trained on Wi-Fi RSSI features.
\end{itemize}

\subsection{Computational Complexity}
We analyze the computational complexity of LiGen with respect to fingerprint dimension $d$, model width $H$, hidden layers $L$, dataset sizes (real $N$, augmented $N_{\text{aug}}$), batch size $B$, and epochs $E$.
\paragraph{Localization MLP.}
For an $L$-layer MLP with width $H$ mapping $\mathbb{R}^d \to \mathbb{R}^2$, the per-sample forward cost is
$$C_{\text{MLP, fwd}} = O(dH + (L-1)H^2 + 2H).$$
Per-epoch training time scales as
$$T_{\text{MLP, epoch}} = O\left(\frac{N + N_{\text{aug}}}{B}\right) \cdot C_{\text{MLP, fwd}}.$$
Test-time inference costs $C_{\text{MLP, fwd}}$ per sample. With $d=10$, the $H^2$ term dominates for large $H$.
\paragraph{PointGAN (Generator + Discriminator)}
Generator and discriminator forward costs are
\[
\begin{aligned}
C_g &= O\big(zH + (L_g - 1)H^2 + dH\big) \\
C_d &= O\big(dH + (L_d - 1)H^2 + H\big)
\end{aligned}
\]

with noise dimension $z$. Per-iteration complexity is
$$T_{\text{PointGAN, iter}} = O(B) \cdot (2C_d + C_g),$$
and per-epoch time
$$T_{\text{PointGAN, epoch}} = O\left(\frac{N}{B}\right) \cdot (2C_d + C_g).$$
Coordinate conditioning adds $O(H)$ to the first layer.
\paragraph{FreeGAN with Weak Localization Model (WLM)}
FreeGAN omits conditioning, matching PointGAN costs up to constants. Pseudo-labeling $N_{\text{free}}$ samples costs
$$T_{\text{WLM, label}} = O\left(\frac{N_{\text{free}}}{B}\right) \cdot C_{\text{WLM, fwd}},$$
with $C_{\text{WLM, fwd}} \ll C_d$ due to smaller width.
\paragraph{End-to-End Training.}
Total time is
$$T_{\text{total}} = E_{\text{GAN}} \cdot T_{\text{GAN, epoch}} + E_{\text{MLP}} \cdot T_{\text{MLP, epoch}} + T_{\text{WLM, label}}.$$
The $O(H^2)$ terms dominate since $d$ is small ($d=10$) and $L$ is modest ($L=5$)
\paragraph{Inference Complexity.}
Deployment requires one MLP forward pass per sample:
$$T_{\text{infer}} = O(dH + (L-1)H^2),$$ 
\paragraph{Practical Considerations.}
With $d=10$ and $H$ in hundreds, training is GPU-efficient. End-to-end training averages 7 minutes on a single NVIDIA RTX A6000 GPU.

\section{Experiment}
\begin{figure}[b]
    \centering
    \includegraphics[width=0.48\textwidth]{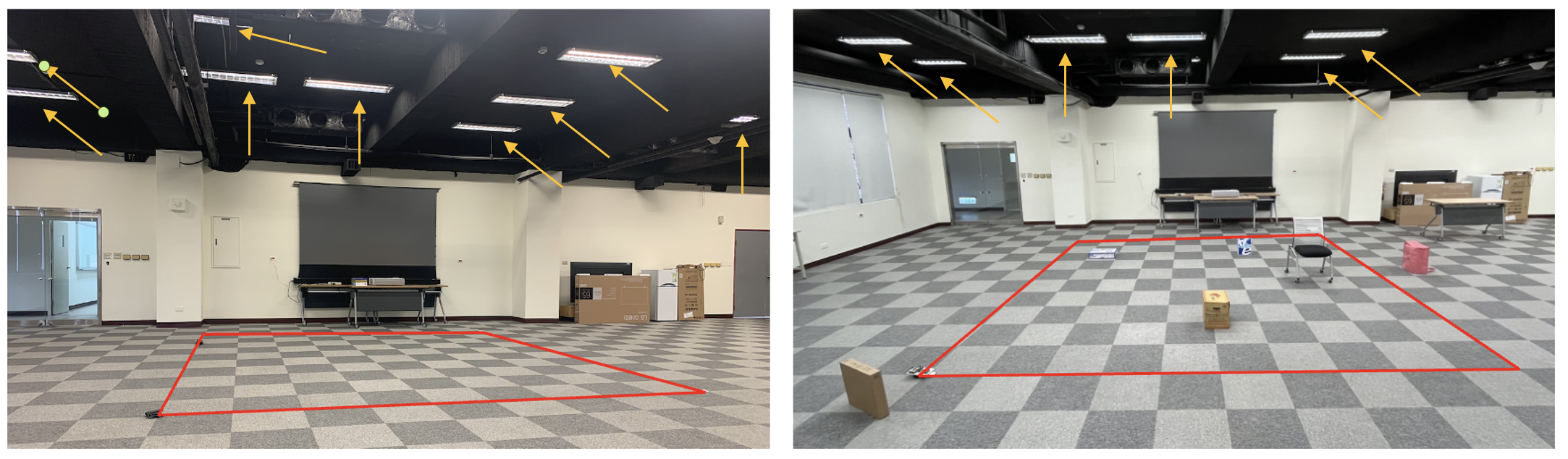}
    \caption{Experimental environments: clean (left) and cluttered with obstacles (right). Yellow arrows indicate ceiling-mounted LED light sources.}
    \label{fig:clean-vs-clutter}
\end{figure}

\subsection{Data Collection Procedure}

We collected two datasets, \textbf{Spectral-WiFiMix} and \textbf{SpectralRobust}, corresponding to Experiments~1 and~2. Both were gathered in Room~601, CSIE, National Taiwan University. The room is lit by ceiling-mounted LED panels (4000–4500,K) with curtains blocking external sunlight, ensuring controlled lighting conditions.

\subsubsection{Spectral-WiFiMix (7\,m  $\times$ 7\,m, for Experiment~1)}
Collected in a $7\times7$\,m indoor area using an AS7341 spectral sensor~\cite{amsAS7341UserGuide} mounted on a Raspberry Pi 4B cart. The region was divided into a $1$\,m grid, yielding $8\times8=64$ reference points. At each point, the sensor recorded 10 spectral channels and Wi-Fi RSSI from six APs, with 32 consecutive samples at 1\,Hz, for a total of $32\times64=2048$ samples.

\subsubsection{SpectralRobust (5\,m $\times$ 5\,m, for Experiment~2)}
Collected in a $5\times5$\,m indoor area to test robustness under environmental variations. The region was divided into a $0.5$\,m grid, yielding $11\times11=121$ reference points. At each point, 10 spectral samples were collected at 1\,Hz, producing $10\times121=1210$ samples. Two environmental settings were prepared:  
(1) \textit{Clean environment} without obstacles, and  
(2) \textit{Cluttered environment} with chairs, boxes, and bags placed at random positions.  
The cluttered setup is shown in Figure~\ref{fig:clean-vs-clutter}.

\subsubsection{Data Augmentation}
For experiments involving \textbf{FreeGAN}, we generate 50{,}000 unlabeled synthetic fingerprints in a single batch and assign pseudo-labels using the Weak Localization Model (WLM). For experiments involving \textbf{PointGAN}, we generate 100 synthetic fingerprints for each training reference point; for example, in Spectral-WiFiMix with 64 training points, yields $64 \times 100 = 6{,}400$ synthetic samples.

\subsection{Signal Stability Comparison}
Before presenting the main experimental results, we highlight a key motivation for using light spectral fingerprints: their exceptional temporal stability.

To quantify signal stability at a single coordinate, we define the \textit{Normalized Average Standard Deviation (Normalized STD)} as:
$\tfrac{1}{d}\sum_{j=1}^{d}\tfrac{\sigma_j}{\mu_j}$, where \(d\) is the number of channels for light-based measurements or the number of APs for Wi-Fi measurements; \(\mu_j\) and \(\sigma_j\) are the mean and standard deviation of the \(j\)-th channel/AP, respectively, across multiple samples at the same location. This metric measures the average relative fluctuation across all channels or APs, normalized by their mean magnitude.

\begin{figure}[H]
    \centering
    \includegraphics[width=0.5\textwidth]{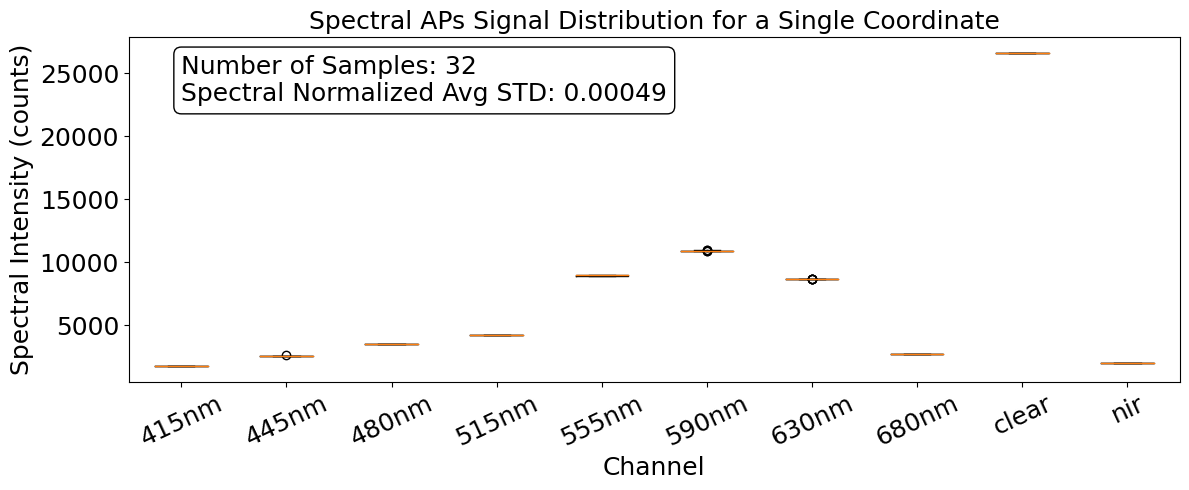}
    \caption{Spectral channel distribution for a single coordinate (32 samples). Normalized Avg. STD: 0.00049}
    \label{fig:light-stability}
\end{figure}

\begin{figure}[H]
    \centering
    \includegraphics[width=0.5\textwidth]{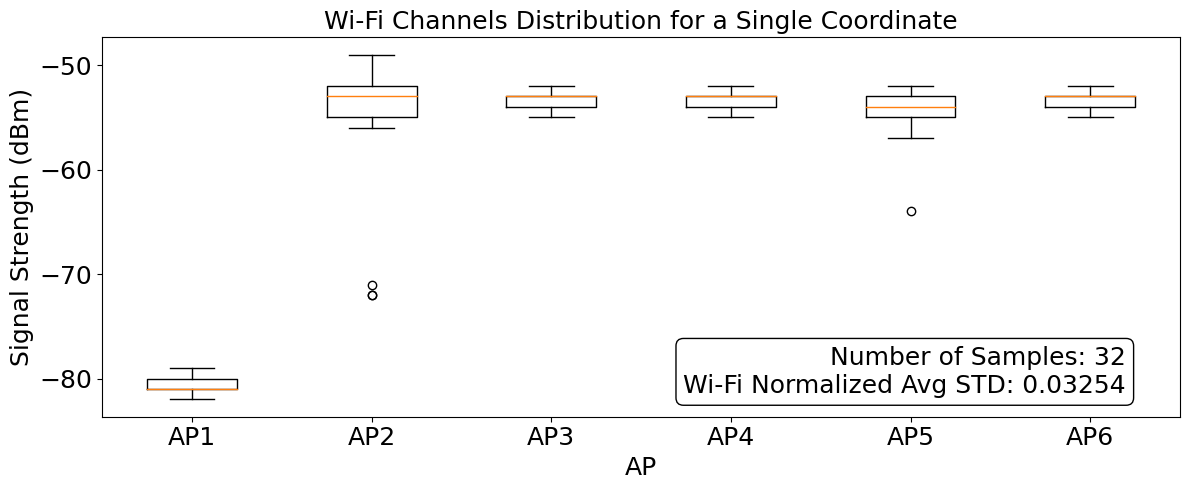}
    \caption{Wi-Fi AP signal strength distribution for a single coordinate (32 samples). Normalized Avg. STD: 0.03254}
    \label{fig:wifi-stability}
\end{figure}

As shown in Figures~\ref{fig:light-stability} and \ref{fig:wifi-stability}. The spectral data exhibits a Normalized STD of 0.00049, significantly lower than the 0.03254 observed for Wi-Fi, demonstrating that light-based fingerprints provide much higher temporal stability.

\section{Experimental Results}
\vspace{-0.1cm}
We evaluate the performance of \textit{LiGen} against the baseline \textit{DataLoc+}~\cite{Hilal2021-DataLoc} under two coordinate-level evaluation settings:

\begin{itemize}
    \item \textbf{Experiment 1: Coordinate-Level Random Split (Spectral vs.\ Wi-Fi).}
    On the \textit{Spectral-WiFiMix} dataset, we allocate $x\%$ of coordinates to training and the rest to testing, per protocols in GAN-based localization~\cite{Chan2024WiFiGANInpainting}. Comparing spectral and Wi-Fi signal performance under same models.
    \item \textbf{Experiment 2: Coordinate-Level Split with Environmental Variation.}
    On \textit{SpectralRobust}, we employ identical partitioning while introducing environmental perturbations, assessing outcomes in clean and cluttered conditions.
\end{itemize}

\begin{figure}[t]
    \centering
    \includegraphics[width=0.95\linewidth]{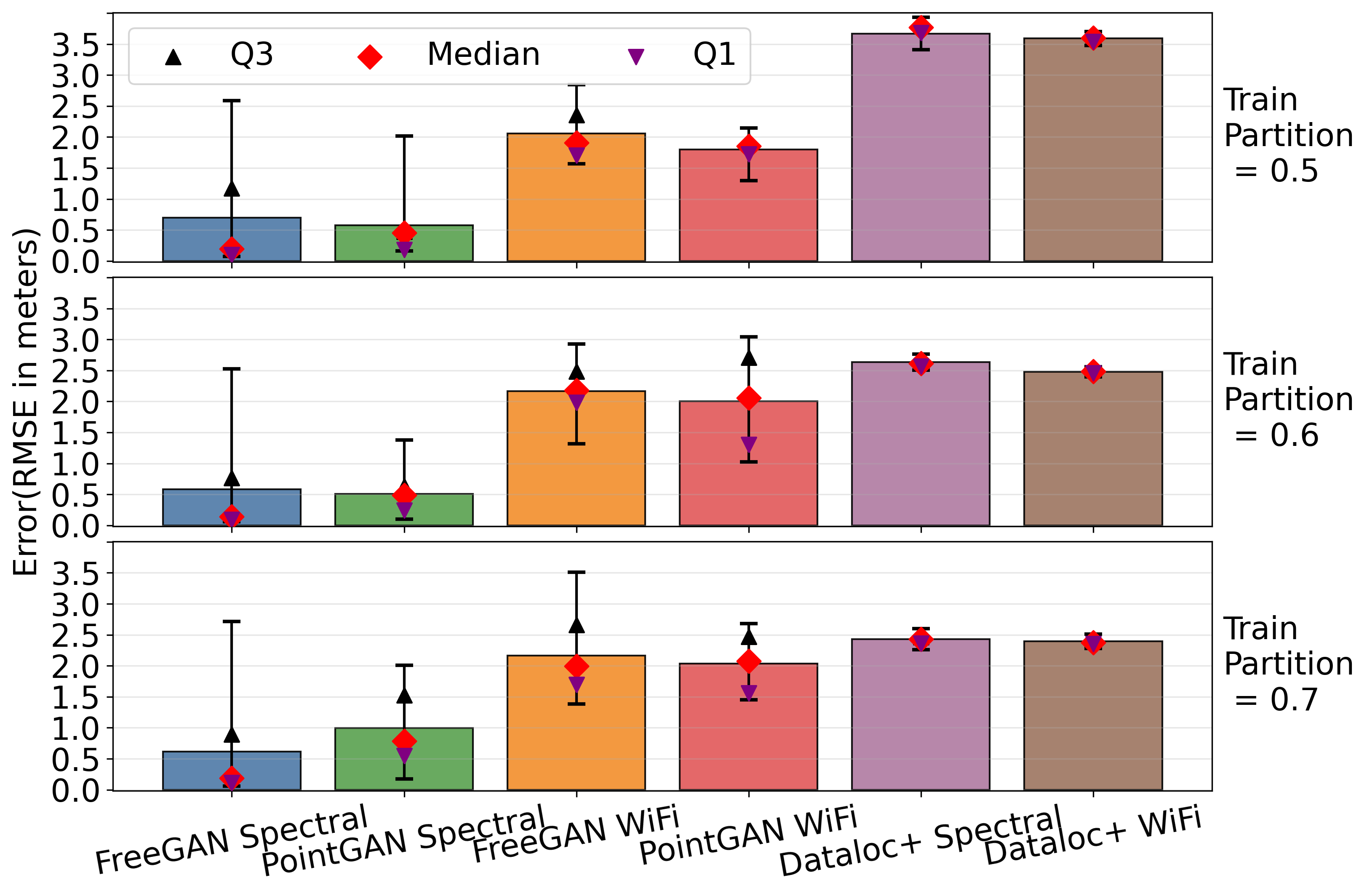}
    \caption{Result of Experiment 1(i): Comparison between Wi-Fi and light signal with data augmentation.}
    \label{fig:experiment1}
\end{figure}
\begin{figure}[b]
    \centering
    \includegraphics[width=0.95\linewidth]{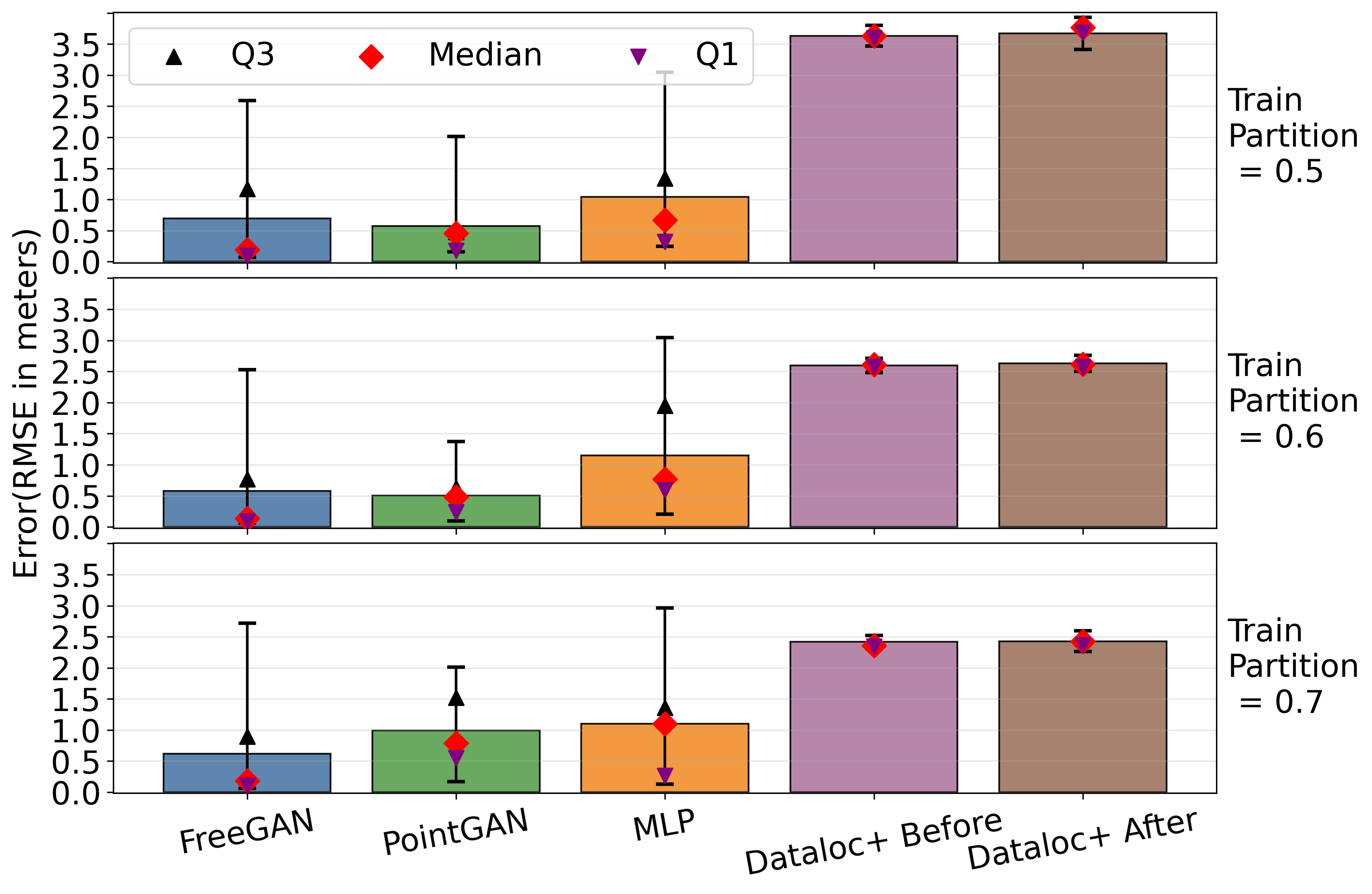}
    \caption{Result of Experiment 1(ii): Performance comparison before and after data augmentation using light signal only.}
    \label{fig:experiment2}
\end{figure}

\vspace{-0.25cm}
\subsection{Experiment 1: Coordinate-Level Random Split (Spectral-WiFiMix)}

In this experiment, we compare spectral and Wi-Fi inputs. Training of \textit{LiGen}'s generative and localization modules, alongside the baseline, excluded data from designated test coordinates. Train on $x \in \{50,60,70\}$ coordinates, with training loss converging around 70\%. RMSE values derive from 20 independent runs with varied random seeds, probing generalization to novel points and resilience to sparse data.

GAN-based models, irrespective of input modality, surpass \textit{DataLoc+} baselines (Figure~\ref{fig:experiment1}). Spectral inputs exhibit substantially lower errors than Wi-Fi counterparts, with FreeGAN on spectral data attaining superior overall accuracy. While maximum errors occasionally elevate, the third quartile ($q_3$) stays well below these peaks across configurations, signaling infrequent outliers and reliable core predictions. This disparity highlights robust sampling's role in enhancing generalization and mitigating tail risks. Performance plateaus after modest data volumes, indicating efficient convergence.

Data augmentation further bolsters spectral inputs. The baseline yields errors exceeding 2.5 m post-augmentation, whereas a vanilla MLP achieves 1--1.5 m (Figure~\ref{fig:experiment2}), underscoring spectral fingerprints' inherent value. PointGAN, FreeGAN curtail averages below 0.6 m, descending to 0.09 m with augmentation and judicious instance selection---affirming GAN augmentation's marked efficacy.

\subsection{Experiment 2: Coordinate-Level Split with Environmental Variation (SpectralRobust)}

\begin{figure}[H]
    \centering
    \includegraphics[width=1.0\linewidth]{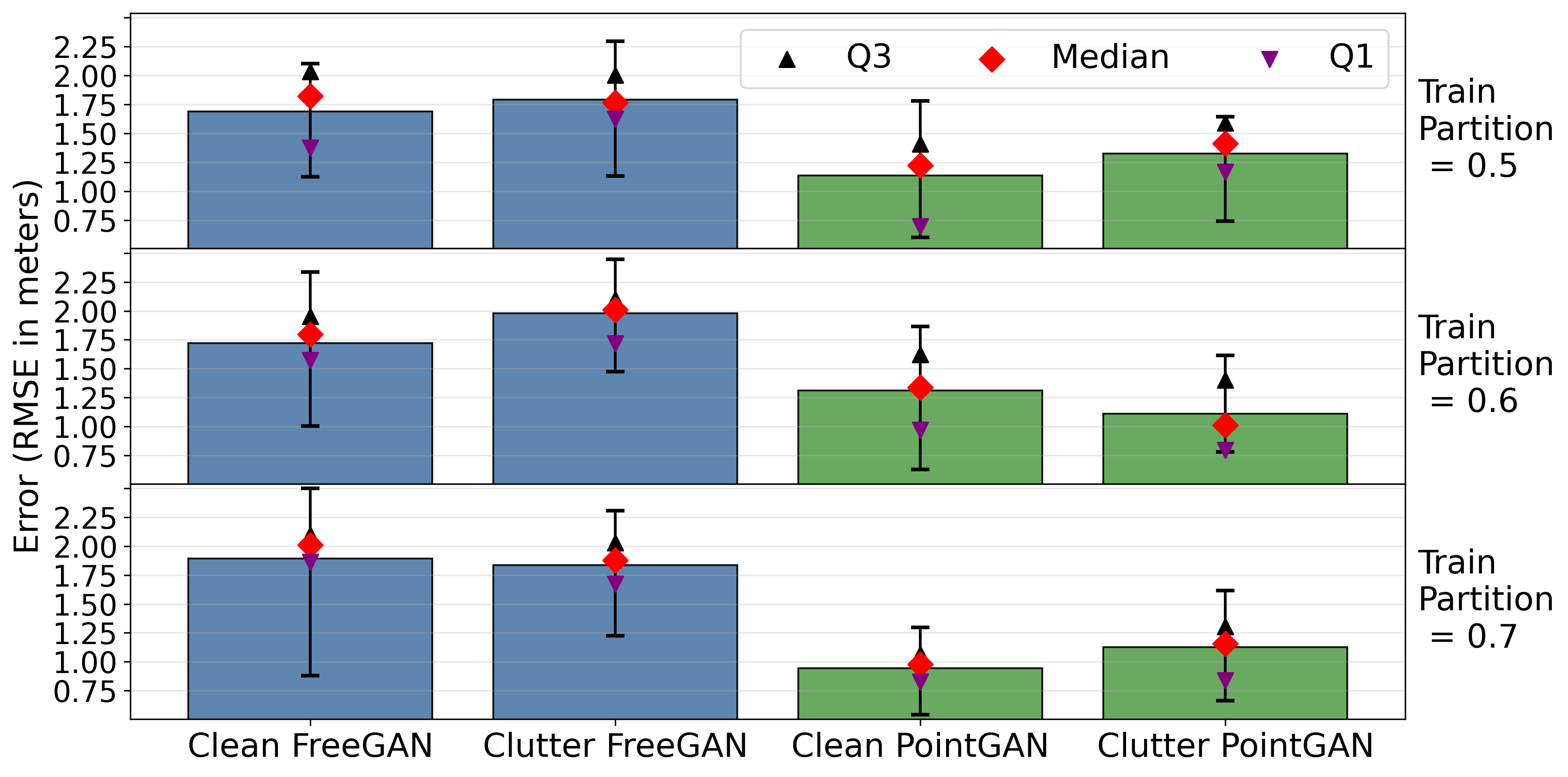}
    \caption{Result of Experiment 2: Comparison of model performance in clean versus cluttered environments.}
    \label{fig:experiment3}
\end{figure}

In SpectralRobust, we assess LiGen in two settings—clean and cluttered—to test robustness under realistic variations. As shown in Figure~\ref{fig:experiment3}, performance remains consistent, with best-case errors differing by  \textless0.6\,m. However, the GAN augmentors diverge: \textit{FreeGAN} averages ~1.75\,m, while \textit{PointGAN} achieves ~1.2\,m. We attribute this gap to augmentation bias: \textit{PointGAN} enforces local geometric structure in spectral fingerprints, preserving coordinate-specific cues important under reflective clutter; \textit{FreeGAN}, optimized for global fidelity with weaker spatial constraints, yields smoother, less diverse samples that underrepresent clutter-induced multipath and specular artifacts, causing residual domain shift. Strengthening \textit{FreeGAN} with spatial consistency losses, physics-informed constraints, or clutter-aware instance selection may narrow the gap while retaining generative flexibility.

\vspace{-0.2cm}
\section{Conclusion}
This work shows that spectral fingerprints markedly improve indoor localization over Wi‑Fi RSSI. LiGen leverages spectral properties and GAN-based augmentation to reach sub‑meter precision, cutting error from 3.20\,m to 0.09\,m. Integrating PointGAN and FreeGAN enriches training with synthetic fingerprints and sustains robust performance in cluttered environments. These results highlight the promise of combining optical fingerprints with generative models to address sparse, variable datasets. Future work will scale LiGen to larger, dynamic spaces and refine GANs to capture complex environmental changes, moving toward better localization.

\bibliographystyle{IEEEtran}
\bibliography{main}

\end{document}